# Analysis of the Ethiopic Twitter Dataset for Abusive Speech in Amharic

**Seid Muhie Yimam**[1], **Abinew Ali Ayele**[1,2], **Chris Biemann**[1],

Language Technology Group, Department of Informatics, MIN Faculty
Universität Hamburg, Germany[1],
Faculty of Computing, Bahir Dar Institute of Technology
Bahir Dar University, Ethiopia[2]
{yimam, ayele, biemann}@informatik.uni-hamburg.de

## Abstract

In this paper, we present an analysis of the first Ethiopic Twitter Dataset for the Amharic language targeted for recognizing abusive speech. The dataset has been collected since 2014 that is written in Fidel script. Since several languages can be written using the Fidel script, we have used the existing Amharic, Tigrinya and Ge'ez corpora to retain only the Amharic tweets. We have analyzed the tweets for abusive speech content with the following targets: Analyze the distribution and tendency of abusive speech content over time and compare the abusive speech content between a Twitter and general reference Amharic corpus.

**Keywords:** abusive speech, hate speech, offensive speech, less-resourced language, Amharic tweet

## ረቂቅ

በዚህ ጽሑፍ ውስጥ የጥላቻ ንግግሮችን ለመለየት ለሚደረጉ ጥናቶች የሚያገለግል የአማርኛ ቋንቋ የመጀመሪያ የትዊተር የውህብ ስብስብ ዳሰሳ ጥናት አቅርበናል። የኢትዮኪክ የትዊተር ማህበራዊ ትስስር የውህብ ስብስብ ከ2014 እ.ኤ.አ ጀምሮ ተሰብስቢል። በኢትዮኪክ ፊደል (ግዕዝ ፊደል) የተጻፉ ትዊቶች ብቻ ተለይተው በአንድ የመረጃ ቋት ውስጥ ተቀምጠዋል። የግዕዝ ፊደልን በመጠቀም የሚፃፉ በርካታ ሌሎች ቋንቋዎችም ስላሉ፣ የአማርኛ ትዊቶችን ብቻ ለመለየት አሁን ላይ የሚገኙ የአማርኛ ፣ የትግርኛ እና የግዕዝ የዕሁፍ ስብስቦችን ተጠቅመናል። የጥላቻ ንግግሮችን ይዘት በተመለከተ ከትዊተር የተገኘውን ዕሁፍ የዳሰስነው በሚከተሉት አቅጣጫዎች ነው፡- 1) የጥላቻ ንግግሮችን ይዘት፣ ስርጭት እና ዝንባሌ ከጊዜ ሂደት ጋር መተንተን፣ 2) በትዊተር የማህበራዊ ትስስር ጽሑፍ እና በአጠቃላይ የአማርኛ ማጣቀሻ የመረጃ ስብስብ መካከል ያለውን የጥላቻ ንግግሮች ይዘት ማነፃፀር።

## 1. Introduction

The emergence of social media creates seamless communication between people and hugely increases the level of information sharing. In the Ethiopian case, people use social media as a primary source of information, and they tend to believe everything from these sources. Recently, we have witnessed a large level of chaos in Ethiopia due to misinformation and abusive language dissemination using social media. The hate speech and fake news dissemination already affected the lives of millions, schools and universities recently closed, business activities heavily hampered due to closure of main roads in the country, the movement of citizens has been seriously hindered, and millions are displaced while hundreds have died (Kiruga, 2019).

It is now a global trend to fight the dissemination of false news and abusive language. Some of the nations have already created regulations that should be compliant with freedom of speech[1] (Levush, 2019).

At the beginning of 2019, the Ethiopian government has drafted legislation[2] against hate speech and hold a series of discussions with different stakeholders, where it is expected to be a law once approved by the parliament before the end of the year.

In this paper, the primary focus is to briefly analyze the Ethiopic Twitter Dataset (ETD) towards abusive speech for Amharic. We hope that this paper, in general, serves as a basis for future research concerning social media contents and, in particular, to study the abusive speech usage and trends in social media for the Amharic language. It further opens a dialogue between technology practitioners, law enforcement parties, and citizens as well on how to deal, regulate and counter attack abusive speech using social media[3].

## 2. Motivation of the Study

The emergence of social media, particularly Facebook and Twitter facilitate the way people communicate in their day-to-day activities. It makes the communication and sharing of information much faster and easier. It brings a friend closer than ever, which otherwise not possible to maintain such links. In the case of Ethiopian social media communication, it is believed that the connection between the larger population of the Diaspora and the friends at home is getting much easier. Furthermore, it has facilitated the transfer of knowledge and technology much simpler and more affordable.

Despite such huge positive influences, social media is bringing its negative consequences to the Ethiopian population than other developing countries (Sibhat, 2018). Social media makes the dissemination of rumors, false information, and hate speech much faster, as a larger portion of the population is already using smartphones for their daily communications.

---

[1] https://www.poynter.org/ifcn/anti-misinformation-actions/
[2] https://bit.ly/2KDSVDx
[3] This paper tries to highlight the coverage of abusive languages on social media content based on a list of keywords collected form limited audiences. We do not yet conduct a proper abusive language analysis and can not also declare a given word, phrase, or sentence as an abusive or not. Moreover, topics discussed are not based on a specific law from the Ethiopian constitution, rather they are based on a general and technological notion that is adopted in the global arena of hate and offensive speech research.

The article by Dibaba (2019) pointed out that the dissemination of hate speech is endangering the democratic rights, jeopardize the long-standing social fabrics and ultimately create political and socio-physiological havoc destabilizing the country. The definition of abusive texts in this paper is confined to the definition of the new draft regulation that is proposed by the Ethiopian government this year.

### 2.1. The New Ethiopian Draft Regulation about Hate Speech

The socio-political crisis that existed since 2016 in Ethiopia, has caused devastating ethnic and sometimes religious-based conflicts. Many people died, displaced from their villages, private and government buildings were also destroyed. The role of hate speech spanning through social media in aggravating these devastating mass conflicts was paramount. It has been noted that hate speech, in the current polarized Ethiopian politics escalates the danger of ethnic and sometimes religious-based mass conflicts by inciting the public (Sibhat, 2018).

In April 2019, the attorney general of the Federal Democratic Republic of Ethiopia has prepared a draft law[4] to tackle hate speech and fake news. In this 5 page draft, which is prepared in Amharic, it describes what defines hate speech and fake news in more general terms. Particularly, it defines *hate speech* as a speech that targets an individual, group or community based on religion, race or color, gender or physical appearance, immigration or origin, and language that intentionally depicts the target as evil, demeans, threatens, discriminates, or otherwise evoke violence.

In this regard, hate speech is targeting a certain ethnic or a specific political group and religion that jeopardizes the exercise of human and democratic rights in the country. Moreover, hate speech threatens the peaceful social life, the long-lasting unity of people and even may lead to a massive massacre between ethnic as well as religious groups if not managed by regulation. Therefore the need for a regulation to govern hate speech is very critical and timely (Dibaba, 2019).

However, the draft is criticized as being poorly drafted with profound implications for human rights in general and freedom of expression as well as the right to privacy in particular. The draft is also blamed for confusing social media with the conventional media[5]. It also fails to impose clear criminal responsibility on hatred social media users and many other vague and confusing even unseen scenarios that should seriously be considered (Abraha, 2019).

### 3. Dataset Collection

### 3.1. General Reference Corpora

While our main purpose is to analyze the content of the ETD for abusive languages in Amharic, we also collect and analyze general reference corpora (GRC) mainly 1) used to train language models for language identification tasks, and 2) to examine the distribution of the selected keywords for abusive language. Even though there are more than 10 Ethiopian languages that use the Ethiopic script (the Fidel) for their writing system, we have obtained a textual dataset only for three languages, namely Amharic (GRC-AM), Tigrinya (GRC-TI), and Ge'ez (GRC-GE). The size and description of these corpora are presented in Section 3.3.

### 3.2. Twitter Dataset

The Ethiopic Twitter Dataset for Amharic (ETD-AM), which is the main focus of analysis in this paper, is collected from mid-August 2014 and continues collecting the tweets written in Fidel script every day. We have collected specifically texts written with Fidel script. Our program runs every day and fetches the tweet, date, time, user location, tweet ID. Until now, around three million tweets have been collected from 154,477 users.

### 3.3. Language Identification and Separation

Since the Fidel script used as writing system for various **Ethiopian** and **Eritrean** languages, such as **Argobba**, **Awngi**, **Blin**, **Chaha**, **Dizin**, **Harari**, **Inor**, **Silt'e**, **Tigre**, **Tigrinya** and **Xamtanga**[6], we have developed a language identification and separation component. There is no publicly available tool to detect and identify texts written in Fidel script into their respective language families (Semitic languages). For the three Ethiopic languages, namely Amharic, Tigrinya, and Ge'ez, there are corpora of sufficient size that can be used to train a model for language detection.

Amharic and Tigrinya are currently used both in academic and daily information propagation (mainly traditional news outlet and social media texts) while Ge'ez is mainly used in the production and dissemination of religious texts by the Ethiopian Orthodox Church (Molla, 2018). We suppose that the ETD we have collected is a mixture of mainly these three languages. To identify the languages of each tweet, we build a language model based on the work of Cavnar and Trenkle (1994), which uses N-gram frequency statistics. For Amharic, we have texts from three sources, 1) web-corpus texts that we have collected at Universität Hamburg using a focused crawler, 2) from the Opus repository[7] (Tiedemann, 2012) where they have more than 300 parallel corpus text, and 3) from the Amharic web corpus (Suchomel and Rychlý, 2016a). For Tigrinya, we use texts from the Opus repository and the Tigrinya web corpus (Suchomel and Rychlý, 2016b). For Ge'ez, we have manually crawled religious books from the Scripture Tools for Every Person (STEP)[8] and from the Lexical Data Repository of the Ge'ez Frontier Foundation[9]. Those tweets other than this stated three languages are categorized as other and trimmed out from our analysis since we do not find available datasets to build the respective language identification model. Table 1 shows the statistics of the three corpora (upper half) and the distribution and statistics of tweets identified into the three languages (lower part) while Table 2 displays the top 5 frequent n-grams.

---

[4] `https://bit.ly/2KDSVDx`
[5] `https://theowp.org/ethiopias-drafted-legislation-against-hate-speech-threatens-journalistic-freedoms/`
[6] `https://www.omniglot.com/writing/ethiopic.htm`
[7] `http://opus.nlpl.eu/`
[8] `https://www.stepbible.org/version.jsp?version=Geez`
[9] `https://github.com/geezorg/data`

| Language | Tokens | Types |
|---|---|---|
| GRC-AM | 46,353,602 | 1,363,192 |
| GRC-TI | 8,512,177 | 339,189 |
| GRC-GE | 316,740 | 42,721 |
| ETD-AM | 26,277,724 | 1,097,986 |
| ETD-TI | 3,152,168 | 309,851 |
| ETD-GE | 385,336 | 52,114 |
| ETD-Other | 195,326 | 19,777 |

Table 1: The number of tokens and types (unique occurrences of tokens) in the ETD and GRC dataset. The suffix *-AM*, *-TI*, and *-GE* stands for *Amharic*, *Tigrinya*, and *Ge'ez* respectively. In the Ethiopic Twitter Dataset, when the text written in the Fidel script can not be identified as either *Amharic*, *Tigrinya*, or *Geez*, it is placed in a separate group as *Other*.

## 4. Abusive Language in ETD-AM

In this section, we will analyze the nature and distribution of abusive texts in Amharic using the ETD based on keywords collected from 5 native speakers. The tweets we use for the analysis are only the Amharic tweets that are identified and filtered by the language model.

In the following sub-section, we will analyze particularly the emergence and proliferation of abusive speech on Twitter. All charts show normalized frequencies in the unit of parts per million (ppm).

### 4.1. Keywords for Abusive speech

In this paper, we adopt the definition of hate and offensive speech based on the work of Davidson et al. (2017). The distinction between hate and offensive speech is always blurry, and we believe that it also depends on the languages, situations or context, and times of the events. We define keywords as hate speech if it fits the definition of the current draft legislation. Otherwise, we categorize the keywords as offensive speech.

We have collected 99 hate speech and 48 offensive speech keywords for the Amharic language from different participants (native speakers)[10]. The participants have collected the keywords from Facebook posts and comments, Twitter tweets and re-tweets, and Youtube comments from popular pages.

### 4.2. Analysis of Abusive Speech in Amharic

Based on the keywords we have collected, we have analyzed the ETD-AM from different aspects. Since the dataset has been collected for 5 years, we first analyze how the keywords are distributed in the dataset. The ETD collected in

---

[10]We select participants who are actively engaging in social media and who are from different fields of study (Political science, Journalism, Engineering, Business administration, and Computational linguistics).

2015 were not correctly stored in our database due to an encoding issue. Hence we do not analyze the dataset for this year.

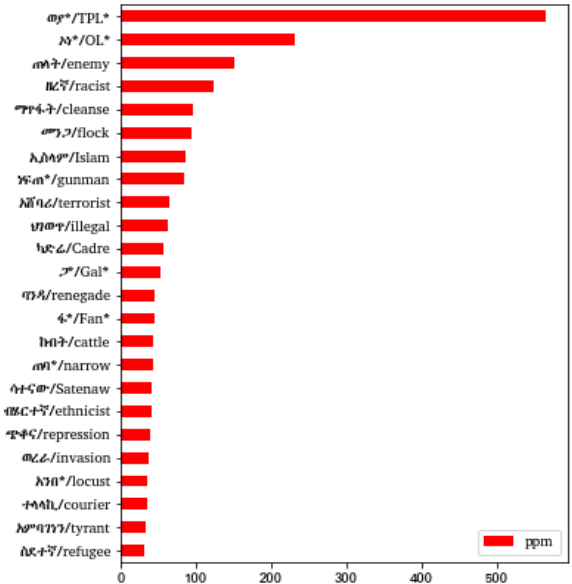

Figure 1: Distribution of hate speech keywords

Figure 1 shows the frequencies of hate speech keywords while Figure 2 shows the frequencies of offensive keywords in the Amharic Twitter dataset. From these figures, we can see that the frequencies of hate speech keywords are very large compared to their offensive counterparts.

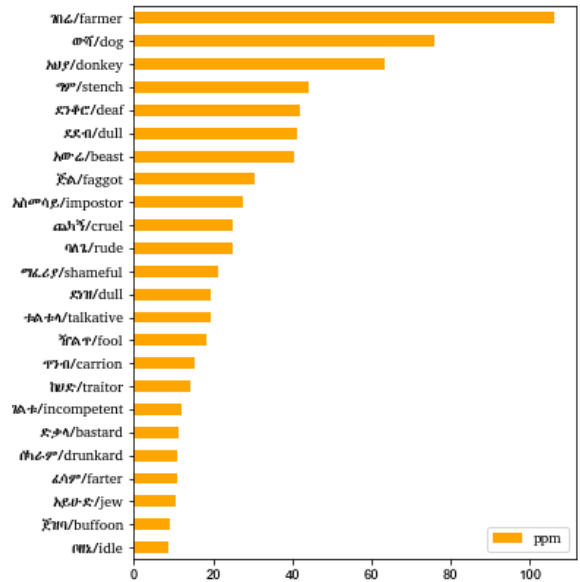

Figure 2: Distribution of offensive speech keywords

From Figure 3, it can be seen that the number of tweets is increasing over time. The same holds for the number of people using Twitter social media are also increasing continuously.

An interesting analysis is observed when we compare the distribution of hate and offensive speech keywords in the ETD-AM and GRC-AM. Even if there are quite a large

| Uni-grams | | Bi-grams | | Tri-Grams | |
|---|---|---|---|---|---|
| Word | Freq. | Phrase | Freq. | Phrase | Freq. |
| ነው/is | 346,965 | አዲስ አበባ/Addis Ababa | 13,538 | ዶ/ር አብይ አህመድ/Dr. Abiy Ahmed | 3,778 |
| ላይ/on | 138,466 | አብይ አህመድ/Abiy Ahmed | 10,954 | ላይክ እና ሼር/like and share | 3,718 |
| እና/and | 125,040 | ብቻ ነው/and only | 9,372 | እርሶም ትኩስ መረጃዎችን/you too hot-news | 2,066 |
| ግን/but | 60,580 | በአዲስ አበባ/by Addis Ababa | 8,262 | ጠ ሚ አቢይ/PM Abiy | 1,963 |
| ሰው/man | 56,502 | የአዲስ አበባ/of Addis Ababa | 8,185 | እንኳን ደስ አለዎት/Congratulations | 1,917 |

Table 2: The most five frequent Ngrams from Amharic tweets

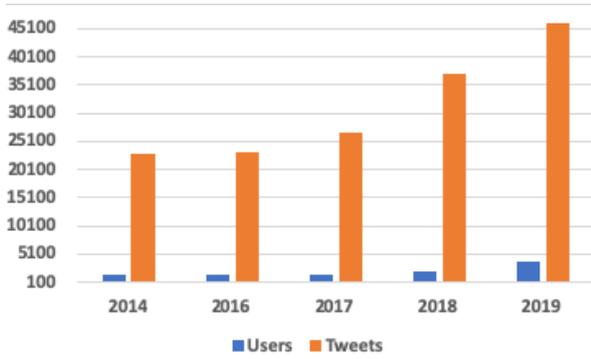

Figure 3: Number of users and Amharic tweets in the ETD per year for the last five years

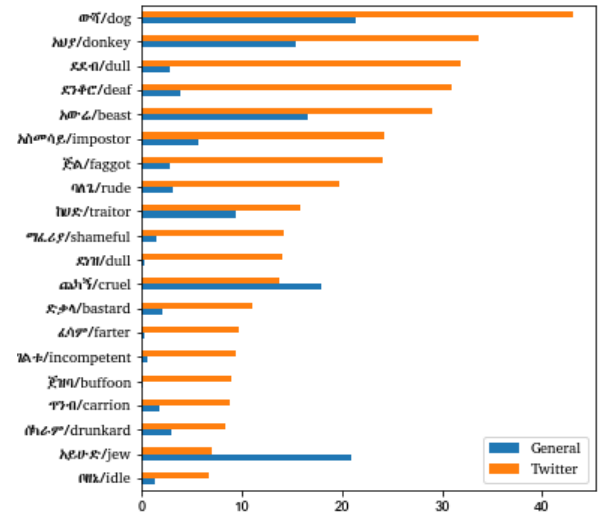

Figure 5: Comparison of offensive speech keywords (y-axis) between the GRC-AM and the ETD-AM, based on their respective ppm (x-axis).

## 5. Conclusion

In this paper, we report the distribution of abusive speech for the Amharic language based on the Ethiopic Twitter Dataset. We have collected around 144 abusive speech keywords from 5 native speakers and categorize them into hate and offensive speech. We then analyze how abusive speech develop over the last five years. In general, the total amount of Amharic Tweets, as well as the number of tweets containing abusive keywords, are increasing over time. The dataset will be used to build automatic abusive language detection systems for Amharic.

## 6. Bibliographical References

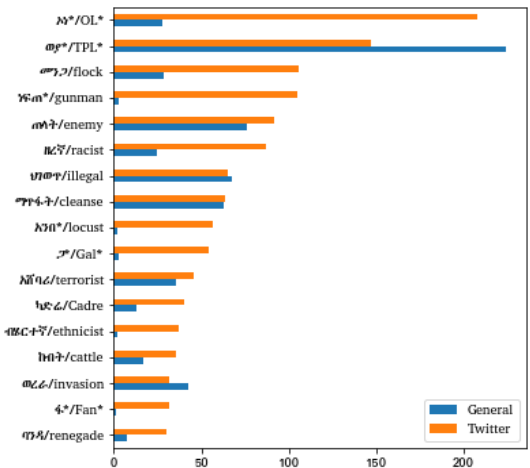

Figure 4: Comparison of hate speech keywords (y-axis) between the GRC-AM and the ETD-AM, based on their respective ppm (x-axis).

number of texts in the general domain, particularly abusive keywords have occurred more often in the ETD-AM than the GRC-AM (See Figure 4 and 5). Keywords that are particularly used conventional news portals such as organization names (example TPLF) are more dominant in the GRC-AM dataset than in the ETD-AM dataset. Whereas, if the organization name is labeled as abusive by the mainstream media (example OLF), the term appears more in the ETD than in the GRC-AM dataset.